%% file: main.tex
\documentclass{article}

\usepackage{microtype}
\usepackage{graphicx}
\usepackage{subfigure}
\usepackage{enumitem}
\usepackage{booktabs} %
\usepackage{multirow}
\usepackage{adjustbox}
\usepackage[most]{tcolorbox}
\usepackage{listings}
\usepackage{makecell}
\usepackage{tikz}
\usetikzlibrary{arrows.meta, positioning}

\usepackage{hyperref}

\usepackage[accepted]{icml2026}

\usepackage{amsmath}
\usepackage{amssymb}
\usepackage{mathtools}
\usepackage{amsthm}
\usepackage{wrapfig}

\usepackage[capitalize,noabbrev]{cleveref}

\usepackage{comment}

\usepackage{titlesec}

\titlespacing*{\section}{0pt}{0.5ex}{0.5ex}

\theoremstyle{plain}

\theoremstyle{definition}

\theoremstyle{remark}

\usepackage[textsize=tiny]{todonotes}

\newcommand{\ourtitle}{%
    On the Resilience of Text-to-Video Diffusion Models to Hardware Faults}
\newcommand{\ourtitleshort}{
    \ourtitle}
\newcommand{\topic}[1]{\noindent\textbf{#1}}

\icmltitlerunning{\ourtitleshort}

\begin{document}

\twocolumn[
\icmltitle{\ourtitle}

\icmlsetsymbol{equal}{*}

\begin{icmlauthorlist}
\icmlauthor{Zachary Coalson}{osu}
\icmlauthor{A M Aahad}{osu}
\icmlauthor{Stella Doehring}{osu}
\icmlauthor{Zane Ma}{osu}
\icmlauthor{Sanghyun Hong}{osu}
\end{icmlauthorlist}

\icmlaffiliation{osu}{Oregon State University, Corvallis, OR, USA}

\icmlcorrespondingauthor{Zachary Coalson}{coalsonz@oregonstate.edu}
\icmlkeywords{Machine Learning, ICML}

\vskip 0.3in]

\printAffiliationsAndNotice{}  %

\input{sections/0_abstract}

\input{sections/1_intro}
\input{sections/2_prelim}
\input{sections/3_method}
\input{sections/4_results}
\input{sections/5_conclusion}

\input{sections/8_ack}

{
    \bibliography{bib/this_paper}
    \bibliographystyle{icml2026}
}

\newpage
\appendix
\onecolumn
\input{sections/9_appendix}

\end{document}

%% file: sections/0_abstract.tex
\begin{abstract}

We present the first systematic study of 
the resilience of text-to-video (T2V) diffusion models 
under random hardware-level faults.
While T2V models are widely used for automated video generation 
due to their ability to produce high-quality, temporally coherent, and realistic videos, 
their iterative denoising process and spatiotemporal dependencies 
introduce unique failure modes. 
We perform an extensive fault-injection study 
covering both computational and memory faults 
across three T2V models and a representative benchmark.
Our results show that 
(1) a single fault can degrade overall performance by up to 3.7\%, 
with semantic correctness more affected than perceptual quality; 
(2) memory faults are more damaging than computational faults, 
high-order exponent bits are particularly vulnerable, and the widely-used bfloat16 is more susceptible than alternative formats; and
(3) 7--28\% of faults cause visible artifacts, including semantic changes such as added objects, suggesting that single faults are sufficient to alter output semantics.
Our findings reveal reliability risks in deployed T2V systems and 
motivate further research on improving fault resilience.
Code: \href{https://github.com/ztcoalson/T2V-Resilience}{https://github.com/ztcoalson/T2V-Resilience}.
\end{abstract}

%% file: sections/1_intro.tex
\section{Introduction}
\label{sec:intro}

Text-to-video (T2V) diffusion models~\cite{zheng2024opensora, ma2025latte, yang2025cogvideox} generate high-quality, coherent videos from text prompts, enabling diverse applications such as physical-world simulation~\cite{videoworldsimulators2024}, educational content generation~\cite{yang2024probabilistic}, automated animation creation~\cite{guo2023animatediff, he2023animate}, and even movie production~\cite{zhu2023moviefactory}.
At the core of these models is a latent Diffusion Transformer (DiT) backbone~\cite{peebles2023scalable}, which uses billions of learned parameters to iteratively denoise latent representations into prompt-aligned videos.
This process incurs substantial compute and memory overhead, driving deployment onto GPU infrastructure as models continue to scale.

Large-scale GPU infrastructure, however, is susceptible to \emph{random bitwise faults}~\cite{oles2024understanding, zhu2025understanding, tiwari2015reliability}: system-level soft errors caused by external factors such as radiation~\cite{abhishek2022radiation} and electromagnetic interference~\cite{mutlu2015rowhammer, lin2025gpuhammer}.
Although error-correcting codes (ECC) can correct single-bit memory errors, multi-bit faults and computational errors often remain uncorrected~\cite{oles2024understanding} and can propagate into the weights or activations of T2V models during inference.

Despite these concerns, the resilience of T2V diffusion models to random bitwise faults remains unexplored.
Prior fault resilience studies have focused primarily on classification networks~\cite{reagen2018ares, li2017understanding, roquet2024cross}, with more recent work extending to large language models~\cite{agarwal2023resilience, sun2025demystifying}.
However, these findings do not directly transfer to T2V: the iterative, diffusion-based inference process, spatiotemporal output dependencies, and absence of discrete correctness criteria all distinguish them from previously studied architectures.

\input{figures/t2v_architecture}

\topic{Contributions.}
In this work, we present the first systematic study of 
the resilience of T2V diffusion models to random bitwise hardware-level faults.
T2V diffusion models are particularly vulnerable to such faults due to three factors:
(1) iterative reuse of backbone parameters, 
which can amplify errors across denoising steps;
(2) spatiotemporal dependencies across frames~\cite{ma2025latte},
where small perturbations may cause motion inconsistencies or degraded coherence; and
(3) the lack of a well-defined reference for output quality,
making fault impacts difficult to quantify.

We design a fault-injection framework that simulates memory and computational faults within the DiT backbone, and evaluate their impact on generation quality, temporal consistency, and semantic alignment using 16 individual and 3 aggregate metrics from a comprehensive T2V benchmark.
We apply this framework to three representative T2V diffusion models, 
injecting more than 300{,}000 faults across 550 prompts.
Our analysis yields the following key findings:
\begin{itemize}[itemsep=0.em, leftmargin=1.1em, topsep=0.0em]
\item Random bitwise faults cause a $\sim$0.3--3.7\% drop in overall performance, while semantic correctness degrades up to 6.6$\times$ more than visual quality.
\item 2-bit memory faults are 2.4--8.6$\times$ more impactful than 1-bit computational faults, due to their persistent impact.
\item The most significant exponent bit is disproportionately vulnerable, while sensitivity is largely uniform across Transformer blocks and layer types.
\item The widely used bfloat16 format is more susceptible to faults than alternative numerical representations.
\item %
7--28\% of faults produce visible semantic alterations or severe distortions%
, suggesting that individual faults can substantially alter output semantics. %
\end{itemize}

%% file: figures/t2v_architecture.tex
\begin{figure*}
\centering
\includegraphics[width=0.9\linewidth]{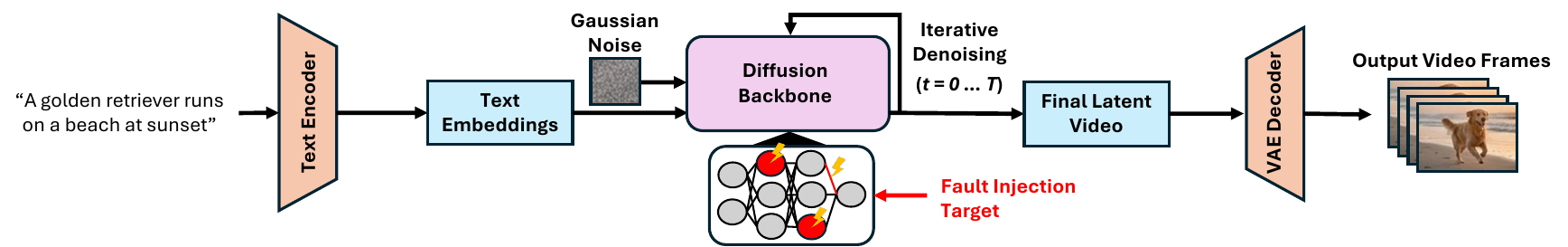}
\caption{\textbf{High-level architecture of a text-to-video diffusion model.}
    The DiT backbone
    operates in latent space and is reused across denoising steps.
    In our study, we assess resilience by injecting random bitwise faults into the backbone's \emph{weights} and \emph{activations}.}
\label{fig:t2v_architecture}
\vspace{-1.6em}
\end{figure*}

%% file: sections/2_prelim.tex
\section{Background and Related Work}
\label{sec:bg}

\subsection{Text-to-Video Diffusion Models}

Text-to-video (T2V) diffusion models extend image diffusion techniques~\cite{saharia2022photorealistic, zhang2024texttoimagediffusionmodelsgenerative} to the spatiotemporal domain, generating coherent video sequences through an iterative denoising process. 
These models build on denoising diffusion probabilistic models~\cite{ho2020denoising} and the latent diffusion framework of \citet{rombach2022ldm}, improving efficiency by operating in a 
compressed latent space rather than pixel space.
Recent architectures implement diffusion backbones with Transformers, using their latent representations to improve the scalability and quality of video synthesis~\cite{zheng2024opensora, ma2025latte, yang2025cogvideox, wan2025wan}.

As shown in Figure~\ref{fig:t2v_architecture}, a typical T2V architecture consists of three primary components: (1) a text encoder %
that produces semantic embeddings of the input prompt, (2) a spatiotemporal Diffusion Transformer (DiT) backbone that denoises latent representations conditioned on these embeddings, and (3) a variational autoencoder (VAE) decoder that maps the final representation back to pixel space.
During inference, the DiT iteratively denoises a sampled latent over $T$ steps, reusing the same backbone weights at each step, before the VAE decodes the result into frames. 
This weight reuse makes the DiT the central component governing generation and our primary target for fault-injection analysis.

\subsection{Fault Resilience of Machine Learning Models}

Fault-resilience studies investigate how machine learning models behave under system-level soft errors
which manifest as bit-flips that corrupt weights or activations.
Early work evaluated the impact of random bitwise faults on classification networks, primarily measuring degradation of top-1 accuracy~\cite{ibrahim2020soft, li2017understanding, hong2019tbd, reagen2018ares}.
Recent studies have extended resilience analysis to Transformer architectures, including image classification~\cite{roquet2024cross} and general language modeling~\cite{agarwal2023resilience}.
Most notably, \citet{sun2025demystifying} conduct a resilience study of large language model inference, evaluating both memory faults and computational faults across multiple tasks %
and models.
However, these findings do not transfer to T2V due to the architectural differences noted in \S\ref{sec:intro}; to our knowledge, no prior work has evaluated the fault resilience of T2V diffusion models.

%% file: sections/3_method.tex
\section{Our Fault Injection Framework}

\subsection{Fault Models}

We study fault models characterizing the origin, target operations, and severity (number of corrupted bits) of hardware faults.
Following recent work~\cite{sun2025demystifying}, we consider two fault sources: (1) GPU memory faults that corrupt model \emph{weights} and (2) computational faults in hardware components (e.g., ALUs) that corrupt \emph{activations}.
For memory faults, we assume a 2-bit fault model---the smallest fault class that escapes SECDED (Single Error Correction, Double Error Detection) correction in modern GPUs~\cite{oles2024understanding, zhu2025understanding, nvidiaECC}---with both flips occurring in the same weight to remain agnostic to ECC implementations. 
For computational faults, we assume a 1-bit fault model, as single-bit faults are widely studied~\cite{sangchoolie2017one} and are not corrected by hardware.

\input{figures/aggregate_results}

\subsection{Fault Injection Methodology}
\label{subsec:fi-method}

\topic{Fault injection setting.} Following prior work~\cite{sun2025demystifying, agarwal2023resilience}, we assume that a T2V diffusion model experiences faults at inference time (rather than during training) to quantify the expected impact on downstream users once the model is deployed. 
We further assume that a fault occurs \emph{once} per inference (i.e., per video generation), since multiple faults within a single inference window are extremely unlikely; this assumption is consistent with prior machine learning resilience studies~\cite{sun2025demystifying, agarwal2023resilience, reagen2018ares, li2017understanding}. 
We inject faults only into the DiT's generation process, excluding the text encoder and VAE, to isolate effects on the latent video generation process. 
Finally, we assume models are deployed using bfloat16~\cite{kalamkar2019bfloat16}, the standard format for modern model deployments~\cite{hfBfloat16, googleBfloat16}; we test alternatives in \S\ref{subsec:datatype-ablation}.

\topic{Fault injection procedure.}
Following prior work~\cite{sun2025demystifying}, we emulate hardware faults by altering values in PyTorch tensors prior to inference.
For \emph{memory faults}, we uniformly sample a DiT weight (by layer and tensor index), uniformly sample two bit positions in [0, 15], and flip the corresponding bits; we then generate one video and revert the weight. 
For \emph{computational faults}, we sample a layer weighted by parameter count (reflecting fault likelihood), then uniformly sample a diffusion timestep, an output neuron, and one bit position in [0, 15]. A PyTorch hook flips the selected bit in the layer's output tensor at the specified timestep; the hook is removed after generating one video.

\subsection{Models and Benchmark}
\label{subsec:models-and-benchmark}

\topic{Models.}
We consider three representative T2V models: (1) OpenSora~\cite{zheng2024opensora}, (2) CogVideoX-2B~\cite{yang2025cogvideox}, and (3) Latte~\cite{ma2025latte}.
All models employ variants of the T5 text encoder~\cite{raffel2020t5}, %
along with model-specific VAEs and %
DiTs, %
whose backbone sizes range from 1B--2B parameters.
These models have been widely adopted in recent T2V studies~\cite{sun2025t2v, adnan2025foresight, guan2025etva, di2025dh, gao2025lemica}, making them well-suited for our analysis.

\topic{Benchmark.}
We use VBench~\cite{huang2024vbench}, 
a comprehensive benchmark for assessing generated video quality.
VBench provides both an evaluation framework with the metrics described below
and a set of prompts spanning 11 evaluation categories (e.g., human action and color).
Following recent work~\cite{adnan2025foresight},
we sample 50 prompts per category, yielding 550 evaluation prompts in total.

\topic{Metrics.}
We evaluate %
generated videos using 16 metrics from VBench~\cite{huang2024vbench},
spanning visual quality and semantic fidelity.
We %
focus on VBench's three aggregate scores: a \emph{Quality Score} (weighted average of 7 metrics, e.g., subject consistency and aesthetic quality), a \emph{Semantic Score} (weighted average of 9 metrics, e.g., object class, human action, and spatial relationship), and a \emph{Total Score} that combines both.
Full metric definitions are in Appendix~\ref{appendix:metric-definitions}.

%% file: figures/aggregate_results.tex
\begin{figure*}[ht]
    \centering
    \includegraphics[width=\linewidth]{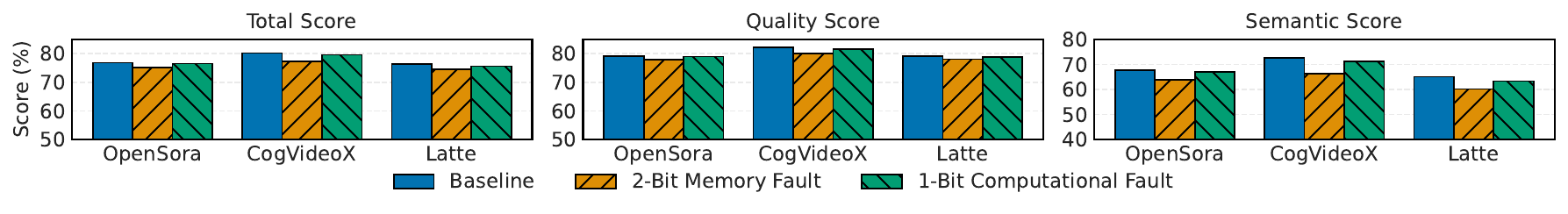}
    \vspace{-1.5em}
    \caption{\textbf{Impact of hardware faults on overall video quality and semantics.}
    Average Total, Quality, and Semantic Scores over 100 trials for the fault-free (baseline), 2-bit memory fault, and 1-bit computational fault models.
    95\% CIs are within $\pm$0.23 across all bars.
}
    \label{fig:aggregate-results}
    \vspace{-1em}
\end{figure*}

%% file: sections/4_results.tex
\input{figures/resiliency_analysis}

\section{Empirical Evaluation}

\subsection{Experimental Setup}
\label{subsec:exp-setup}

\topic{Models and video generation.}
We implement all models using VideoSys~\cite{videosys2024}. %
Following recent work~\cite{adnan2025foresight}, all generated videos are 2 seconds long.
We use each model's default generation settings. %

\topic{Metric computation.}
We conduct 100 trials for each fault model. %
In each trial, a random fault is sampled per video, resulting in \emph{55,000} generations per fault model. 
We then compute VBench metrics per trial and report averages across trials to capture the expected impact of each fault model.

\subsection{Main Resilience Results}
\label{subsec:main-results}

We present results for the three aggregate metrics---Total, Quality, and Semantic Score---in Figure~\ref{fig:aggregate-results}; results for the 16 individual metrics are reported in Appendix~\ref{appendix:individual-metric-results}.
Across fault models and T2V models, the Total Score decreases by 0.3--3.7\%, indicating a modest performance degradation.
Performance drops are similar across T2V models (within $\sim$1\%), suggesting that resilience is largely determined by the DiT backbone rather than model-specific differences.
Examining individual components of performance, we find that video semantics are affected up to 6.6$\times$ more than quality: the Semantic Score drops by 1.1--8.8\%, whereas the Quality Score drops by only 0.2--2.5\%.
This indicates that bit-flips are more likely to alter the attributes and relationships of entities in videos rather than their visual rendering, which may lower the barrier to exploitation via targeted \emph{bit-flip attacks}~\cite{rakin2019BFA, coalson2025prisonbreak}.%

Consistent with \citet{sun2025demystifying}, 2-bit memory faults are 2.4–8.6$\times$ more impactful than 1-bit computational faults (1.4–8.8\% vs. 0.2–2.8\% drops). 
This gap likely arises because memory faults persist across all denoising steps, whereas computational faults affect a single operation; weight faults also influence many more downstream computations through matrix multiplication~\cite{sun2025demystifying}. 

\subsection{Resilience Analysis}
\label{subsec:resiliency-analysis}

\topic{Model-level components.}
We analyze the sensitivity of specific model components under the 2-bit memory fault model: bit positions, Transformer block indices, and layer types. 
For each fault, we record the percentage change across all 16 individual metrics, averaged by component; for bit location, we group by the \emph{maximum} bit position. 
Figure~\ref{fig:resiliency-analysis} shows results for Latte; full results are in Appendix~\ref{appendix:extra-resiliency-analysis}.

Consistent with prior work~\cite{sun2025demystifying, agarwal2023resilience}, we find that the most significant exponent bit (bit 14) is disproportionately vulnerable, causing an average performance drop of 46.3\%, compared to at most 3.4\% for all other bit positions. %
In contrast, vulnerability is largely uniform across other components: performance drops range from 4.8--7.3\% across Transformer blocks and are nearly identical between MLP (6.4\%) and attention (6.3\%) layers. The uniformity across blocks is consistent with findings on other Transformer architectures~\cite{agarwal2023resilience}, though it contrasts with earlier CNN studies that found earlier layers more vulnerable~\cite{li2017understanding}. 

\topic{Diffusion timestep.}
We next measure the resilience of each diffusion timestep, i.e., the step at which the fault is injected. 
As weight faults are always injected prior to inference, we instead consider the 1-bit computational fault model.
For each time step (normalized to $[0,1]$ for consistency across models), we aggregate all faults injected during that step and record the average percentage change in all individual metrics.
Results for all three models are shown in Figure~\ref{fig:timestep-analysis}.

\input{figures/timestep_analysis}

Computational faults appear to have a relatively uniform impact across diffusion timesteps, with metric changes ranging from -3.6\% to +2.6\% and no clear trend as timesteps increase.
While we may expect faults at earlier timesteps to matter more, since the perturbed latent passes through more of the remaining denoising process and thus has a greater opportunity to be amplified, we observe no clear dependence on timestep.
Taken together with the bit-position results above, this suggests that fault magnitude is the primary driver of degradation, rather than its location or timing.

\subsection{Impact of Numerical Representation}
\label{subsec:datatype-ablation}

\input{figures/dtype_ablation}

We examine how numerical representation affects resilience by comparing bfloat16, float16, and float32 under the 2-bit memory fault model on CogVideoX; for float32, we sample bits in [0, 31]. 
Results are shown in Figure~\ref{fig:datatype-ablation}. 
We find that bfloat16 is the most vulnerable format, with its Total Score dropping by 3.1\%, compared to 1.8\% for float32 and 1.2\% for float16. 
This is likely because bfloat16 allocates the most bits to the exponent (8/16) while supporting the same range as float32, making large perturbations more probable.

\subsection{Qualitative Analysis}
\label{subsec:qualitative-analysis}

\input{figures/taxonomy_examples}

We qualitatively analyze fault-induced visual changes by taxonomizing them into three categories: \textbf{Minor} (only minor or imperceptible changes from the original), \textbf{Major} (at least one substantial difference, e.g., a changed background), and \textbf{Distorted} (significant corruption, e.g., pure static). Figure~\ref{fig:taxonomy-examples} illustrates each category under 2-bit memory faults: the major fault adds a tree and alters ground tiling, while the distorted fault produces random static across all frames.
To measure the prevalence of these outcomes, we further randomly select 50 fault-injected videos per T2V model–fault model pair (300 total) and manually assign each to a category. Figure~\ref{fig:qualitative-analysis} shows the distribution by fault model.

\input{figures/qualitative_analysis}

We find that while most faults result in only minor changes (72--93\%), a nontrivial fraction causes major alterations (2--17\%) or even complete distortion of the generated video (5--11\%).
2-bit memory faults produce substantially more non-minor changes than 1-bit computational faults (28\% vs. 7\%), suggesting that persistent corruption of the diffusion process is necessary to produce noticeable visual changes.
These results contrast with the modest quantitative degradation in \S\ref{subsec:main-results}, indicating that standard performance metrics fail to capture the distinct failure modes introduced by faults. 
Moreover, the frequency of major changes shows that a few bit-level modifications are sufficient to substantially alter output semantics, motivating future work on whether such changes can be induced in a targeted manner.

%% file: figures/resiliency_analysis.tex
\begin{figure*}[ht]
    \centering
    \includegraphics[width=\linewidth]{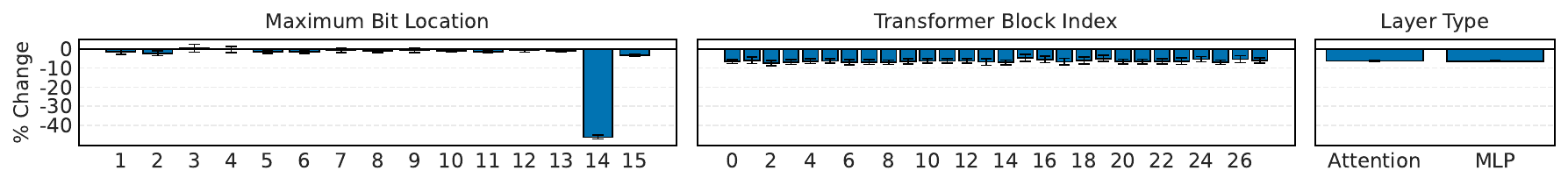}
    \vspace{-1.5em}
    \caption{\textbf{Resilience of model-level components to 2-bit memory faults} for Latte. %
    The average change across individual VBench metrics, categorized by maximum bit location, Transformer block index, and layer type.}
    \label{fig:resiliency-analysis}
    \vspace{-1em}
\end{figure*}

%% file: figures/timestep_analysis.tex
\begin{figure}[ht]
    \centering
    \includegraphics[width=\linewidth]{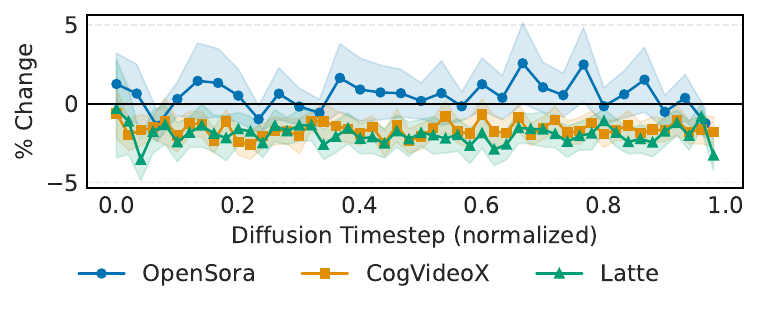}
    \vspace{-1.5em}
    \caption{\textbf{Impact of computational faults across diffusion timesteps}. 
    The average change in individual VBench metrics for each (normalized) diffusion step at which faults were injected.}
    \label{fig:timestep-analysis}
    \vspace{-0.5em}
\end{figure}

%% file: figures/dtype_ablation.tex
\begin{figure}[h]
    \centering
    \includegraphics[width=\linewidth]{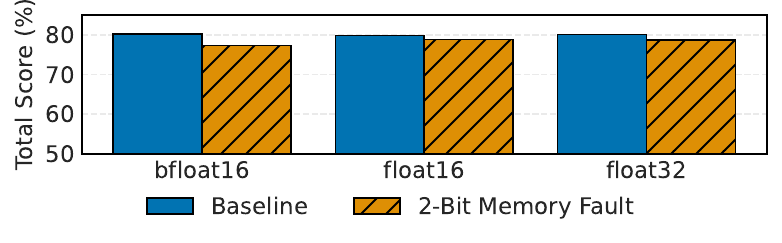}
    \vspace{-2em}
    \caption{\textbf{Impact of numerical representation} for CogVideoX. The average Total Score under the fault-free (baseline) and 2-bit memory fault models.
    95\% CIs are within $\pm$0.08 for all bars.}
    \label{fig:datatype-ablation}
\end{figure}

%% file: figures/taxonomy_examples.tex
\begin{figure}[ht]
    \centering
    \includegraphics[width=\linewidth]{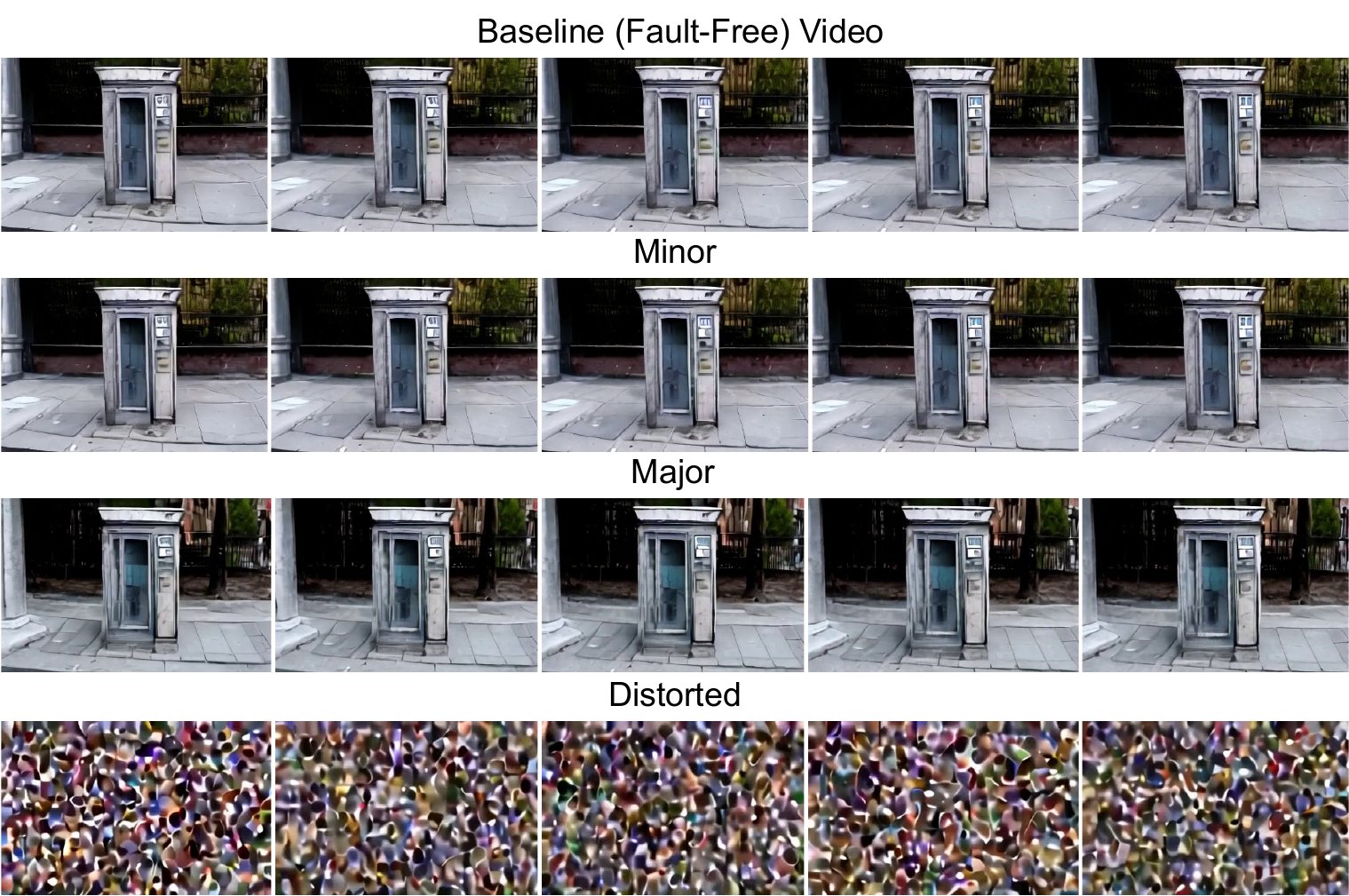}
    \caption{\textbf{Example of qualitative fault outcomes for representative videos} generated from OpenSora. 
    We show 5 evenly spaced frames from each video. 
    Prompt: ``a dilapidated phone booth stood as a relic of a bygone era on the sidewalk, frozen in time.''}
    \label{fig:taxonomy-examples}
\end{figure}

%% file: figures/qualitative_analysis.tex
\begin{figure}[h]
    \centering
    \includegraphics[width=\linewidth]{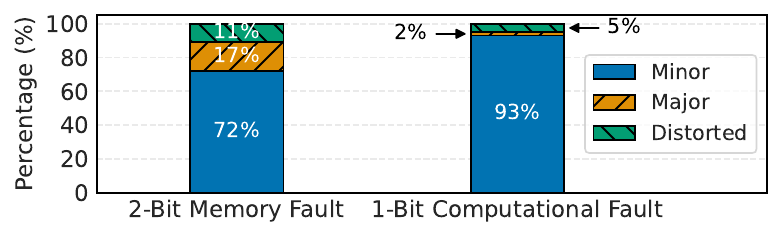}
    \vspace{-1.5em}
    \caption{\textbf{Distribution of qualitative fault outcomes} for the 2-bit memory and 1-bit computational fault models (150 samples each).}
    \label{fig:qualitative-analysis}
\end{figure}

%% file: sections/5_conclusion.tex
\section{Discussion}

\topic{Potential countermeasures.}
Many machine learning defenses against hardware faults exist; we discuss those that appear most promising based on our findings or that are intrinsic to diffusion-based generation.
Since high-order exponent bits account for the majority of degradation (\S\ref{subsec:resiliency-analysis}), defenses should prioritize protecting them.
General redundancy (e.g., full ECC or triplication) corrects such faults but is costly at the scale of billion-parameter DiT backbones, motivating selective protection of only the most vulnerable bits.
Several works have proposed such schemes for other DNN architectures~\cite{catalan2025exploiting, xie2025breaking, fuengfusin2023efficient, zheng2025save}, and adapting them to T2V diffusion models is a promising direction for future work.
An alternative is to correct faults on-the-fly: because the most damaging flips produce large-magnitude perturbations to otherwise bounded weights, range-based clipping~\cite{chen2021low, sun2025ft2, sha2024global} can suppress out-of-distribution activations and mitigate their impact, at the cost of modest compute overhead and the risk of clipping legitimate in-distribution values.
Such methods typically require careful, architecture-specific tuning (e.g., \cite{sun2025ft2}), implying comparable effort would be needed for T2V.
Finally, as faults perturb the iterative denoising trajectory, stochastic sampling methods that contract accumulated errors, such as restart sampling~\cite{xu2023restart}, may dampen their impact at inference.

\topic{Limitations and future work.}
Our study considers two fault models, 2-bit memory and 1-bit computational faults, chosen to reflect uncorrected faults in modern GPUs (\S\ref{subsec:fi-method}); however, other fault models are also practical~\cite{dos2022characterizing, beigi2023systematic} and merit investigation.
We further focus on random faults, as is standard in resilience studies~\cite{sun2025demystifying, agarwal2023resilience, roquet2024cross}. While this characterizes expected-case behavior under naturally occurring faults, several of our findings---particularly the frequency of major semantic alterations (\S\ref{subsec:qualitative-analysis})---suggest faults could be exploited adversarially; confirming whether such effects can be induced in a \emph{targeted} manner is a key open direction.
Finally, our quantitative conclusions rely on VBench metrics, which may not fully capture the impact of faults: as our qualitative analysis (\S\ref{subsec:qualitative-analysis}) shows, the scores understate visible failure modes, motivating metrics better suited to quantifying fault impact.

\section{Conclusion}
\label{sec:conclusion}

As T2V diffusion models are increasingly deployed on large-scale GPU infrastructure, understanding their resilience to hardware faults is critical to ensuring their reliability.
To this end, we present the first systematic fault-injection study of T2V diffusion model inference, evaluating the impact of memory and computational faults on three representative models across 16 video-quality and semantic-correctness metrics.
Our results show that while aggregate performance degradation is modest (at most 3.7\%), this obscures a more complicated picture: semantic correctness is affected more than visual quality, memory faults are significantly more damaging than computational faults, and up to 28\% of faults produce visible alterations---such as added objects or distortion---that standard metrics fail to capture.
These findings demonstrate that T2V diffusion models are \emph{not} robust to random bitwise faults, and that their failure modes differ from those observed in previously studied architectures.

To facilitate further study of T2V resilience and the evaluation of future defenses, we open-source our fault-injection framework at \href{https://github.com/ztcoalson/T2V-Resilience}{https://github.com/ztcoalson/T2V-Resilience}.

%% file: sections/8_ack.tex
\section*{Acknowledgment}
\label{sec:ack}

We thank the anonymous reviewers for their valuable feedback.
This work in part is supported by 
the Samsung Strategic Alliance for Research and Technology (START) program.
Any opinions, findings, and conclusions or recommendations expressed here are those of the authors and do not necessarily reflect the views of the funding agencies.

%% file: sections/9_appendix.tex
\section{Detailed Experimental Setup}
\label{appendix:detailed-setup}

We use Python 3.10.19 with PyTorch 2.2.2 for all experiments, and the official VBench repository\footnote{\href{https://github.com/Vchitect/VBench}{https://github.com/Vchitect/VBench}} to evaluate videos.
We run these experiments on a system with a 48-core Intel Xeon Processor,
768GB of memory and 8 NVIDIA A40 GPUs.

\section{Individual Metric Results}
\label{appendix:individual-metric-results}

\input{figures/individual_results}

We examine changes in performance across the 16 individual metrics computed by VBench; results are shown in Figure~\ref{fig:individual-results}.
Resilience varies substantially across metrics: metrics such as Subject and Background Consistency are minimally affected, with changes ranging from -0.3\% to +1.4\%, whereas others, such as Multiple Objects, Human Action, and Object Class, suffer much larger degradations of 14.2--23.7\%.
Consistent with our aggregate findings, quality metrics are generally more resilient (changes of -8.5\% to +4.5\%) than semantic metrics (changes of -23.7\% to +11.2\%), though it varies considerably within each category.
We attribute these differences to the types of visual perturbations induced by faults.
Specifically, our qualitative analysis (\S\ref{subsec:qualitative-analysis}) reveals that non-imperceptible fault-induced outcomes typically manifest as either semantic alterations (e.g., an added object) or complete distortion (e.g., a black screen).
Such changes likely compromise semantic correctness more than visual quality.

\section{Full Resilience Analysis Results}
\label{appendix:extra-resiliency-analysis}

\input{figures/resiliency_analysis_opensora}

\input{figures/resiliency_analysis_cogvideox}

Here, we complement the resilience analysis on Latte in \S\ref{subsec:resiliency-analysis} with additional results for OpenSora and CogVideoX, shown in Figures~\ref{fig:resiliency-analysis-opensora} and \ref{fig:resiliency-analysis-cogvideox}, respectively.
The results are largely consistent with Latte: the most significant exponent bit is disproportionately vulnerable, while vulnerability is largely uniform across Transformer blocks and layer types.
However, OpenSora exhibits two noticeable discrepancies.
First, all bit positions except the most significant exponent bit show a \emph{positive} average change in individual metrics (up to +3.9\%).
We attribute this primarily to OpenSora exhibiting substantially higher per-trial variance than the other models, as reflected in its wider confidence intervals, which makes the sign of these averages less reliable as an indicator of a systematic effect.
Second, unlike CogVideoX and Latte, attention layers are moderately more resilient than MLPs, with an average change of -0.7\% compared to -2.7\%, suggesting that MLP layers in this model are slightly more sensitive to random weight-level perturbations.

\section{VBench Metric Definitions}
\label{appendix:metric-definitions}

We provide brief definitions of all VBench metrics and refer readers to the original work~\cite{huang2024vbench} for full details:
\begin{itemize}[itemsep=0.em, leftmargin=1.1em, topsep=0.0em]
    \item \textbf{Subject Consistency:} Measures whether subjects (e.g., a person, vehicle, or animal) maintain a consistent appearance across all frames of the video.
    \item \textbf{Background Consistency:} Evaluates the temporal stability of the background, i.e., that the environment remains coherent and does not change unexpectedly over time.
    \item \textbf{Temporal Flickering:} Quantifies high-frequency visual instability between adjacent frames, capturing undesired flicker or jitter artifacts in static regions.
    \item \textbf{Motion Smoothness:} Assesses whether object and camera motions are temporally smooth and physically plausible, without abrupt or unnatural transitions.
    \item \textbf{Dynamic Degree:} Measures the overall amount of motion present in the video, capturing whether the generated content exhibits meaningful dynamics rather than static.
    \item \textbf{Aesthetic Quality:} Evaluates the perceptual appeal of individual frames, including photographic composition, color harmony, and artistic quality.
    \item \textbf{Imaging Quality:} Assesses low-level fidelity of frames, focusing on distortions such as blur, noise, or exposure.
    \item \textbf{Object Class:} Measures whether the generated video correctly depicts the objects specified in the prompt.
    \item \textbf{Multiple Objects:} Evaluates the model's ability to generate and correctly compose multiple distinct objects within the same scene as described by the prompt.
    \item \textbf{Human Action:} Assesses whether human subjects in the video perform the actions indicated in the prompt.
    \item \textbf{Color:} Measures the accuracy with which object colors in the generated video match prompt specifications.
    \item \textbf{Spatial Relationship:} Evaluates whether the spatial relationships between objects (e.g., left/right, above/below) are consistent with the prompt description.
    \item \textbf{Scene:} Measures whether the overall scene (e.g., ocean, city, forest) aligns with the scene described in the prompt.
    \item \textbf{Appearance Style:} Assesses consistency between the visual appearance of frames and the requested artistic or visual style (e.g., watercolor, oil painting).
    \item \textbf{Temporal Style:} Evaluates whether temporal characteristics such as camera motion or cinematic style align with the prompt's temporal style description.
    \item \textbf{Overall Consistency:} Measures holistic alignment between the generated video and the text prompt, capturing both semantic correctness and stylistic coherence.
\end{itemize}
All metrics are computed independently for each generated video and normalized per dimension using empirical minimum and maximum reference values, yielding scores in $[0,1]$ (which we report as percentages).
In addition, we define the three aggregate metrics that summarize overall generation quality across the benchmark:
\begin{itemize}[itemsep=0.em, leftmargin=1.1em, topsep=0.0em]
    \item \textbf{Quality Score:} Aggregates overall video quality as a weighted average of the \emph{first seven} metrics listed above.
    \item \textbf{Semantic Score:} Aggregates prompt-level correctness as a weighted average of the \emph{remaining nine} above metrics.
    \item \textbf{Total Score:} Combines quality and semantic correctness as a weighted average of the Quality and Semantic Scores, placing a $4\times$ greater emphasis on quality.
\end{itemize}
These metrics enable both coarse- and fine-grained analysis of how faults impact diverse aspects of video generation.

%% file: figures/individual_results.tex
\begin{figure*}[!ht]
    \centering
    \includegraphics[width=\linewidth]{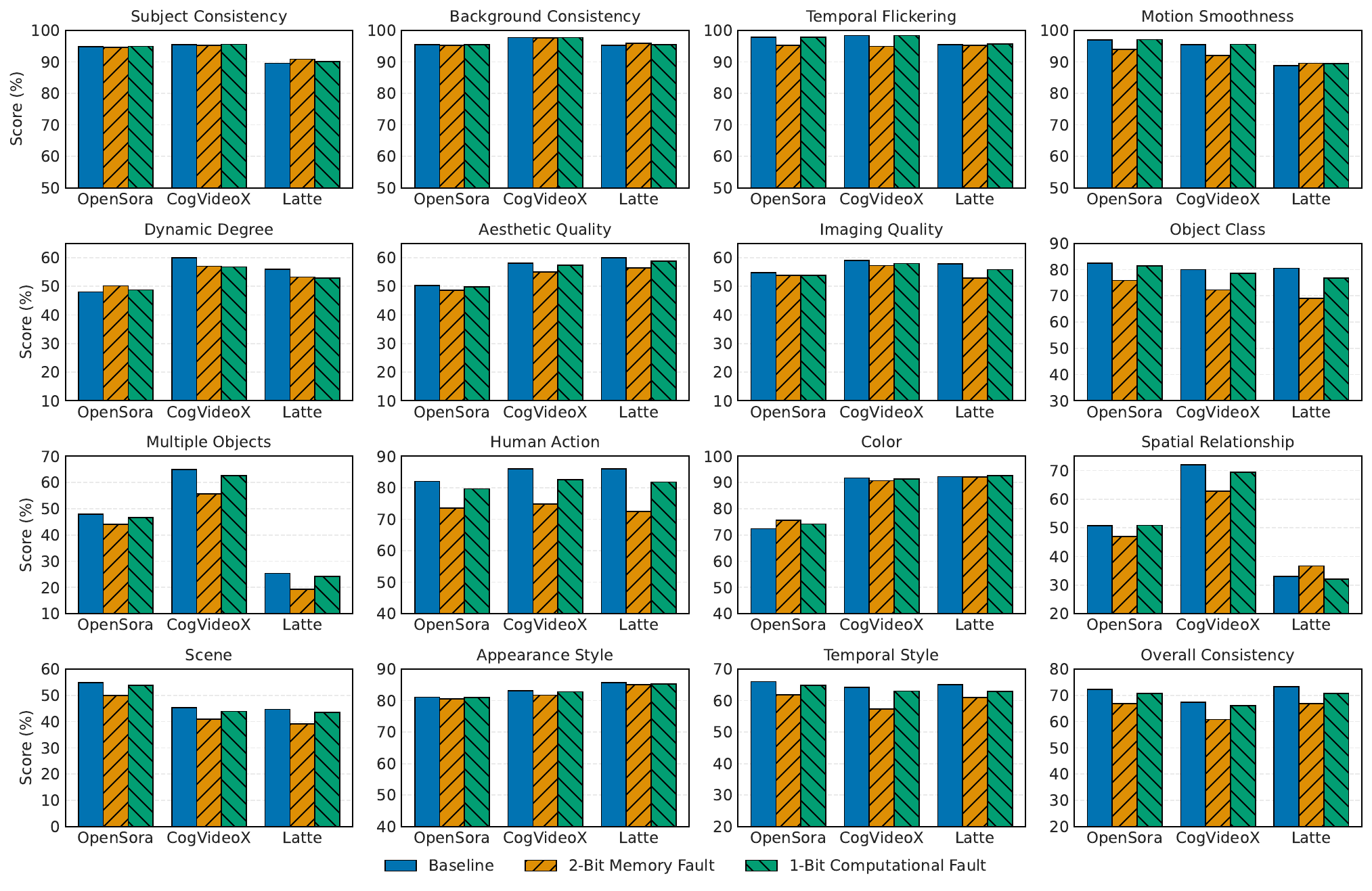}
    \caption{\textbf{Impact of hardware faults on individual video quality and semantics metrics.}
    For each metric, we report the average score over 100 trials for the fault-free (baseline), 2-bit memory fault, and 1-bit computational fault models. 95\% CIs are within $\pm$0.95.}
    \label{fig:individual-results}
\end{figure*}

%% file: figures/resiliency_analysis_opensora.tex
\begin{figure*}[ht]
    \centering
    \includegraphics[width=\linewidth]{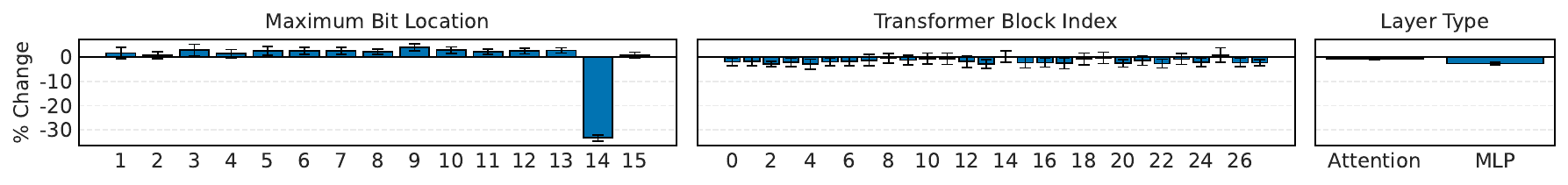}
    \vspace{-1.5em}
    \caption{\textbf{Resilience of model-level components to 2-bit memory faults} for OpenSora. %
    The average change across individual VBench metrics, categorized by maximum bit location, Transformer block index, and layer type.}
    \label{fig:resiliency-analysis-opensora}
    \vspace{-1em}
\end{figure*}

%% file: figures/resiliency_analysis_cogvideox.tex
\begin{figure*}[ht]
    \centering
    \includegraphics[width=\linewidth]{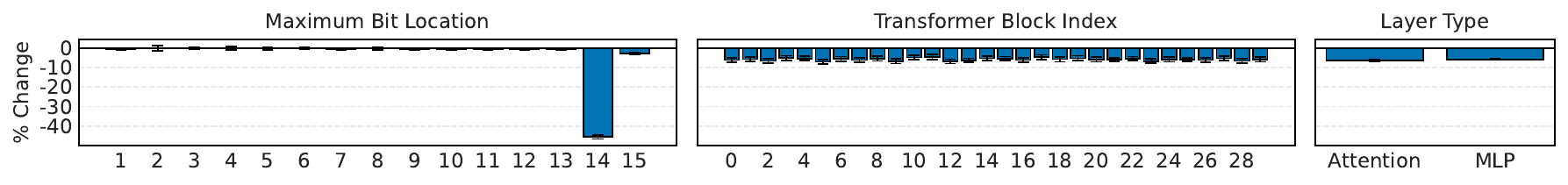}
    \vspace{-1.5em}
    \caption{\textbf{Resilience of model-level components to 2-bit memory faults} for CogVideoX. %
    The average change across individual VBench metrics, categorized by maximum bit location, Transformer block index, and layer type.}
    \label{fig:resiliency-analysis-cogvideox}
\end{figure*}